\documentclass[sigconf]{acmart}

\PassOptionsToPackage{protrusion, expansion}{microtype} 

\usepackage{amsmath, amsthm}
    
\usepackage{graphicx, xcolor} 
\usepackage{hyperref, url}
\usepackage{adjustbox}
\usepackage{booktabs, multirow}
\usepackage{caption}
\usepackage{subcaption}
\usepackage{enumitem}
\usepackage{lipsum}
\usepackage{setspace}
\usepackage{algorithm, algpseudocode}
    \definecolor{BLUE}{rgb}{.0, .2, .6}
    \definecolor{BLUEalt}{HTML}{1e50a2}
    \definecolor{RED}{HTML}{c9171e}
    \algrenewcommand{\alglinenumber}[1]{{\scriptsize\bfseries\ttfamily\color{RED}#1}}

\usepackage{footnote}
\makesavenoteenv{tabular}
\makesavenoteenv{table}

\usepackage[acronym]{glossaries}
\makeglossaries
 
\newglossaryentry{fclayer}{
    name={\texttt{fc}-layer},
    description={fully connected layer}
}
 
\newglossaryentry{szfull}{
    name=\textsc{SZ lossy compression},
    description={SZ lossy compression}
}

\newglossaryentry{deepsz}{
    name=\textsc{DeepSZ},
    description={DeepSZ}
}

\newglossaryentry{dataArray}{
    name=\textsc{data array},
    description={}
}

\newglossaryentry{indexArray}{
    name=\textsc{index array},
    description={}
}
 
\newacronym{gcd}{GCD}{Greatest Common Divisor}
\newacronym{lcm}{LCM}{Least Common Multiple}

\usepackage[zerostyle=d]{newtxtt}

\let\oldTextcolor\textcolor
\renewcommand{\textcolor}[2]{\oldTextcolor{#1}{\sffamily #2}}

\copyrightyear{2019}
\acmYear{2019}
\setcopyright{usgovmixed}
\acmConference[HPDC '19]{The 28th International Symposium on High-Performance Parallel and Distributed Computing}{June 22--29, 2019}{Phoenix, AZ, USA} \acmBooktitle{The 28th International Symposium on High-Performance Parallel and Distributed Computing (HPDC '19), June 22--29, 2019, Phoenix, AZ, USA} \acmPrice{15.00}
\acmDOI{10.1145/3307681.3326608} \acmISBN{978-1-4503-6670-0/19/06}

\begin{document}
\title{\textsc{DeepSZ}: A Novel Framework to Compress Deep Neural Networks by Using Error-Bounded Lossy Compression}



\author{Sian Jin}
\affiliation{%
  \institution{The University of Alabama}
  \city{Tuscaloosa}
  \state{AL}
  \country{USA}
}
\email{sjin6@crimson.ua.edu}

\author{Sheng Di}
\affiliation{%
  \institution{Argonne National Laboratory}
  \city{Lemont}
  \state{IL}
  \country{USA}
}
\email{sdi1@anl.gov}

\author{Xin Liang}
\affiliation{%
  \institution{University of California, Riverside}
  \city{Riverside}
  \state{CA}
  \country{USA}
}
\email{xlian007@ucr.edu}

\author{Jiannan Tian}
\affiliation{%
  \institution{The University of Alabama}
  \city{Tuscaloosa}
  \state{AL}
  \country{USA}
}
\email{jtian10@crimson.ua.edu}

\author{Dingwen Tao}
\authornote{Corresponding author: Dingwen Tao, Department of Computer Science, The University of Alabama, Tuscaloosa, AL 35487, USA.}
\affiliation{%
  \institution{The University of Alabama}
  \city{Tuscaloosa}
  \state{AL}
  \country{USA}
}
\email{tao@cs.ua.edu}

\author{Franck Cappello}
\affiliation{%
  \institution{Argonne National Laboratory}
  \city{Lemont}
  \state{IL}
  \country{USA}
}
\email{cappello@mcs.anl.gov}

\begin{abstract}
Today's deep neural networks (DNNs) are becoming deeper and wider because of increasing demand on the analysis quality and more and more complex applications to resolve. The wide and deep DNNs, however, require large amounts of resources (such as memory, storage, and I/O), significantly restricting their utilization on  resource-constrained platforms. Although some DNN simplification methods (such as weight quantization) have been proposed to address this issue, they suffer from either low compression ratios or high compression errors, which may introduce an expensive fine-tuning overhead (i.e., a costly retraining process for the target inference accuracy). In this paper, we propose \textit{DeepSZ: an accuracy-loss expected neural network compression framework}, which involves four key steps: \textit{network pruning}, \textit{error bound assessment}, \textit{optimization for error bound configuration}, and \textit{compressed model generation}, featuring a high compression ratio and low encoding time.
The contribution is threefold. (1) We develop an adaptive approach to select the feasible error bounds for each layer. (2) We build a model to estimate the overall loss of inference accuracy based on the inference accuracy degradation caused by individual decompressed layers. (3) We develop an efficient optimization algorithm to determine the best-fit configuration of error bounds in order to maximize the compression ratio under the user-set inference accuracy constraint.
Experiments show that 
DeepSZ can compress AlexNet and VGG-16 on the ImageNet dataset by a compression ratio of 46$\times$ and 116$\times$, respectively, and compress LeNet-300-100 and LeNet-5 on the MNIST dataset by a compression ratio of 57$\times$ and 56$\times$, respectively, with only up to 0.3\% loss of inference accuracy. Compared with other state-of-the-art methods, DeepSZ can improve the compression ratio by up to 1.43$\times$, the DNN encoding performance by up to 4.0$\times$ with four V100 GPUs, and the decoding performance by up to 6.2$\times$.
\end{abstract}



\maketitle

\setlength{\textfloatsep}{6pt}

\section{Introduction}

Deep neural networks (DNNs) have  rapidly evolved to the state-of-the-art technique for many artificial intelligence tasks in various science and technology areas, for instance, image and vision recognition \cite{simonyan2014very}, recommender systems \cite{wang2015collaborative}, nature language processing \cite{collobert2008unified}, and time series classification \cite{zheng2014time}. 
DNNs contain millions of parameters in an unparalleled representation, which is efficient for modeling complexity nonlinearities. Thus,  using either deeper or larger DNNs can be an effective way to improve data analysis. As pointed by Wang et al.~\cite{wang2018superneurons}, the deep learning community has been acknowledging that increasing the scales of DNNs can improve the inference accuracy of image recognition tasks. A 9-layer AlexNet \cite{krizhevsky2012imagenet}, for example, proposed by Krizhevsky et al., won the 2012 ILSVRC (ImageNet Large-Scale Visual Recognition Challenge) \cite{ISVRC} with a top-5 accuracy of 83\%. In 2014 ILSVRC, a 22-layer GoogLeNet \cite{szegedy2015going} proposed by Szegedy et al. further improved the record of top-5 accuracy to 93.3\%. He et al. proposed a 152-layer ResNet \cite{he2016deep}, which refreshed the record to 96.43\% in 2015 ILSVRC. This trend suggests that the networks will go larger in the future.

The ever-increasing growth of networks is bringing more and more challenges to resource-limited platforms, such as mobile phones and wireless sensors. For instance, a typical use case is to train DNNs in the cloud using high-performance accelerators, such as graphic processing units (GPUs) or field-programmable gate arrays (FPGAs), and distribute the trained DNN models to edge devices for inferences \cite{teerapittayanon2017distributed}. According to the data released by GSMA \cite{GSMA}, 0.8 billion users will be still using 2G networks (with a theoretical maximum transfer speed of 1 Mbit/s) by 2020. Consequently, one practical challenge is to deliver multiple latest DNN models (i.e., an order of tens to hundreds of megabytes for each model) from cloud to edge devices through bandwidth-limited networks.
An issue also arises in a wireless sensor network with DNN models, which is usually deployed in a region of interest for tracking objects or abnormal detection. Since large DNN models must be stored in external DRAM (e.g., embedded flash-based storage) and frequently fetched for inferences, they consume large amounts of energy \cite{han2015deep}. However, sensor nodes are usually equipped with small batteries and it is unfeasible to recharge them or deploy new nodes in many cases. Therefore, how to maintain/extend the lifetime of sensor networks with large DNNs becomes challenging \cite{wang2008compression}.
Compressing neural networks provides an effective way to reduce the burden of these problems. Most  approaches, however, have focused on simplification methods, such as network pruning \cite{han2015learning} and quantization \cite{han2015deep}, which suffer from limited compression quality. A straightforward idea is to leverage existing lossy compression and encoding techniques \cite{reagen2017weightless} to significantly improve the ratio for compressing DNNs. The existing compression strategies applied on DNNs, however, do not have error expected features, which may greatly distort the data, leading to expensive fine-tuning overhead (i.e., extra, costly retraining process).  

In this paper, we propose \gls{deepsz}: a lossy compression framework for DNNs. \gls{deepsz} is composed of four key steps:  \textit{network pruning}, \textit{error bound assessment}, \textit{optimization of error bound configuration}, and \textit{compressed model generation}. Unlike  traditional compression methods used on DNNs, we perform error-bounded lossy compression on the pruned weights, an approach that can significantly reduce the data size while restricting the loss of inference accuracy. 
Specifically, we adapt the SZ lossy compression framework developed by us previously \cite{di2016fast, tao2017significantly, liangerror} to fit the context of DNN compression. In this compression framework, each data point's value would be predicted based on its neighboring data points by an adaptive, best-fit prediction method (either a Lorenzo predictor or linear regression-based predictor \cite{liangerror}). Then, each floating-point weight value would be converted to an integer number by a linear-scaling quantization based on the difference between the real value and predicted value and a specific error bound. Huffman encoding or other lossless compression such as Zstd \cite{zstd} and Blosc \cite{blosc} would be applied to significantly reduce the data size thereafter. SZ can get a much higher compression ratio on the compression of nonzero weights than other state-of-the-art compressors such as ZFP \cite{zfp} can, especially because of efficient linear-scaling quantization, which contrasts with the simple vector quantization applied to the original weights in other related work \cite{gong2014compressing, han2015deep}. Moreover, our  SZ compressor can control errors in more sophisticated ways, such as relative error bound and peak-signal-to-noise ratio (PSNR).

Designing an efficient lossy compression framework for DNNs raises two important challenging issues to resolve. (1) How can we determine an appropriate error bound for each layer in the neural network? Specifically, we have to explore a feasible range of error bounds for each layer, under which the lossy compression should still get a high inference accuracy for users.
(2) How can we maximize the overall compression ratio regarding different layers in the DNN under user-specified loss of inference accuracy? 
Considering the heterogeneous and diverse data features across multiple layers, we have to explore the best-fit error bounds on the compression of different layers. 
A straightforward idea is to traverse all the possible error-bound combinations on different layers,
which would definitely lead to an extremely high time-complexity. To address this issue, we develop a dynamic strategy to efficiently determine the best-fit error bound for each layer. 

The contributions of this work are summarized as follows.
\begin{itemize}
    \item We propose a novel, accuracy-loss expected framework, called \gls{deepsz}, by applying our previously developed SZ error-bounded lossy compression to compress DNNs. To the best of our knowledge, this is the first attempt to do so. 
    \item We propose an adaptive method to select the feasible range of error bounds for each layer. We also develop an effective model to estimate the overall inference accuracy loss based on the forward-propagation results with individual layers reconstructed from the error-bounded lossy compressor. 
    \item We develop a dynamic algorithm to optimize the combined configuration regarding different layers' error bounds, significantly reducing the overall size of the neural network. In addition, because of our careful design of the accurate error control, our solution also effectively eliminates the costly retraining overhead that was generally introduced by other DNN compression methods \cite{han2015deep, reagen2017weightless}.
    \item We compare  \gls{deepsz} with two other state-of-the-art  works (i.e., Deep Compression and Weightless) based on four well-known neural networks. Evaluation results demonstrate that the compression ratio of \gls{deepsz} is 1.21$\times$$\sim$1.43$\times$  higher than that of the other two approaches. Experiments show that \gls{deepsz} can obtain 1.8$\times$$\sim$4.0$\times$ encoding performance improvement on four Nvidia Tesla V100 GPUs and 4.5$\times$$\sim$6.2$\times$ decoding performance improvement on Intel Xeon Gold 6148 CPU over the second-best approach.
\end{itemize}

The rest of the paper is organized as follows. 
In Section~\ref{sec:background}, we discuss the background and motivation of our research. In Section \ref{sec:design}, we describe the design methodologies of the \gls{deepsz} framework in detail. In Section~\ref{sec:comparison}, we provide a detailed analysis and comparison of \gls{deepsz} and two other state-of-the-art approaches. In Section~\ref{sec:evaluation}, we present the evaluation results on four well-known DNNs with multiple GPUs. In Section~\ref{sec:related}, we discuss related work. In Section~\ref{sec:conclusion}, we summarize our conclusions and present ideas for future work.
\section{Background and Motivation}
\label{sec:background}

In this section, we present some background information about  neural networks and lossy compression for floating-point data and discuss our research motivation.

\subsection{Neural Networks}

Neural networks have been widely studied and used in recent years and have produced dramatic improvements in many scientific and engineering aspects, such as computer vision \cite{rowley1998neural} and natural language processing \cite{collobert2008unified}. Each neural network is composed of multiprocessing layers. Among the various kinds of layers, convolutional layers and fully connected layers (denoted by \glspl{fclayer}) have contributed the most to the recent progress in the deep learning community, especially in vision-related tasks such as image classification and object detection. One convolutional layer consists of a set of filters that slide in the input dataset and perform convolution with the signal in the sliding window. \glspl{fclayer} are connected by a dense weight matrix and forward the signals by a matrix-matrix multiplication. The filters in the convolutional layers and the weight matrices in the \glspl{fclayer} dominate the storage space of the neural networks, which will become larger as the networks become deeper or wider. In neural networks, the forward pass refers to the calculation process, which traverses from the first layer to the last layer. The backward pass refers to the process to update the weights by stochastic gradient descent, which traverses from the last layer backward to the first layer. During the training period, both   forward and backward passes are performed, whereas only the forward pass is performed for testing. In the following discussion, \textsc{test} refers to the forward pass process on the test dataset for generating the inference accuracy of the neural network.

\begin{table}[]
    \footnotesize
    \caption{Architectures of example nerual networks.}
    \label{tab:NeuralNetworks}
    \begin{tabular}{@{}rrrrr@{}}
\toprule
\begin{tabular}{@{}r@{}} \bfseries Neural\\\bfseries Networks\end{tabular} & 
    \begin{tabular}{@{}r@{}} \bfseries LeNet-\\\bfseries 300-100\end{tabular} &
        \bfseries LeNet-5 & 
            \bfseries AlexNet & 
                \bfseries VGG-16  \\
\midrule
\footnotesize\bfseries conv layers &\footnotesize 0 &\footnotesize 3 &\footnotesize 5 &\footnotesize 13      \\
\footnotesize\bfseries \glspl{fclayer}  &\footnotesize 3 &\footnotesize 2 &\footnotesize 3 &\footnotesize 3       \\
\footnotesize\bfseries \texttt{ip1}/\texttt{fc6} & \footnotesize $300\times784$ & \footnotesize $500\times800$ & \footnotesize $4096\times9216$ & \footnotesize $4096\times25088$  \\
\footnotesize\bfseries \texttt{ip2}/\texttt{fc7} & \footnotesize $100\times300$ & \footnotesize $10\times500$ & \footnotesize $4096\times4096$ & \footnotesize $4096\times4096$\\
\footnotesize\bfseries \texttt{ip3}/\texttt{fc8} & \footnotesize $10\times100$ & - & \footnotesize $1000\times4096$ & \footnotesize $1000\times4096$\\
\footnotesize\bfseries {conv fwd time} & \footnotesize 0 ms &\footnotesize 0.5 ms &\footnotesize 116.5 ms &\footnotesize 149.8 ms\\
\footnotesize\bfseries \texttt{fc} fwd time &\footnotesize 0.30 ms &\footnotesize 0.12 ms &\footnotesize 2.5 ms &\footnotesize 1.7 ms\\
\footnotesize\bfseries total size &\footnotesize 1.1 MB &\footnotesize 1.7 MB &\footnotesize 243.9 MB &\footnotesize 553.4 MB \\
\footnotesize\bfseries \glspl{fclayer}' size (\%) &\footnotesize 100\% &\footnotesize 95.3\%  &\footnotesize 96.1\%  &\footnotesize 89.4\%  \\
\bottomrule
\end{tabular}
\end{table}

According to prior studies \cite{alwani2016fused}, although convolutional layers occupy most of the computation time ($\sim$95\%) because of the expensive convolution operations, they take up little storage space ($\sim$5\%). On the other hand, \glspl{fclayer} require large storage space ($\sim$95\%) because of the large dense matrices, while  consuming little computation time ($\sim$5\%). This phenomenon is also verified in our experiments. For  demonstration purposes, we present the breakdown of storage and computational overhead for four well-known networks. As shown in Table~\ref{tab:NeuralNetworks}, the \glspl{fclayer} take the majority of the networks' storage space (i.e., 89.4\% $\sim$ 96.1\%) in all  three cases; however, they have much lower computational cost (i.e., about 1\% $\sim$ 2\% for VGG-16 and AlexNet and 20\% for LeNet-5) than the convolutional layers do. Hence, we are motivated to leverage lossy compression techniques in order to trade computation time for storage space on the \glspl{fclayer} in resource-constrained scenarios,\; our aim is to significantly reduce the storage requirements of neural networks while introducing little computation overhead.

\subsection{Lossy Compression for Floating-Point Data}
Floating-point data compression has been studied for decades. The data compressors can be split into two categories: lossless and lossy. Lossless compressors such as GZIP \cite{deutsch1996gzip}, FPZIP \cite{lindstrom2006fast}, and BlosC \cite{alted2017blosc} cannot significantly reduce the floating-point data size because of the significant randomness of the ending mantissa bits. The compression ratios of lossless compression are generally limited to 2:1, according to recent studies \cite{son2014data, lindstromerror}.

Many  lossy compressors supporting floating-point data were proposed originally for  visualization. Hence, many lossy compressors employ the techniques directly inherited from lossy compression of images, such as variations of wavelet transforms, coefficient prioritization, and vector quantization. Lossy compressors for image processing are designed and optimized considering human perception, such as JPEG2000 \cite{taubman2012jpeg2000}. While such compressors may be adequate for scientific visualization, they do not provide pointwise error controls on demand. For example, most lossy compressors designed for visualization do not provide control of a global upper bound on the compression error (the maximum compression error, or $L_{inf}$ norm of the compression error). 

A new generation of lossy compression techniques for floating-point data has been developed recently.  SZ, ZFP, 
and MGARD 
\footnote{Some compressors such as ISABELA \cite{isabela} were designed with pointwise error control, but tests \cite{di2016fast} have shown that the maximum error could be much larger than the user-set error bound.} are three typical  error-bounded compressors.  
SZ \cite{di2016fast, tao2017significantly, liangerror} predicts each data point's value by its neighboring data points in a multidimensional space with an adaptive predictor (using either a Lorenzo predictor \cite{ibarria2003out} or linear regression \cite{liangerror}). Next, it performs an error-controlled linear-scaling quantization to convert all floating-point values to an array of integer numbers. And then it performs a customized Huffman coding and lossless compression to shrink the data size significantly. ZFP \cite{zfp} splits the whole dataset into many small blocks and compresses the data in each block separately by four steps:  alignment of exponent, orthogonal transform, fixed-point integer conversion, and bit-plane-based embedded coding.
MGARD uses multigrid methods to compress multidimensional floating-point data  \cite{ainsworth2018multilevel}. 
Many independent studies \cite{di2016fast, tao2017significantly, liangerror, lu2018understanding, tao2017depth} have showed that SZ outperforms the other two compressors in terms of compression ratio, especially on 1-D 
floating-point datasets; note that  the datasets to compress in our case are 1-D floating-point arrays after conversion. 

Today's lossy compression techniques have been used in HPC scientific applications for saving storage space and reducing the I/O cost of saving data. However, \textit{how to effectively and efficiently utilize error-bounded lossy compressors to significantly reduce the neural network size and encoding time, while still maintaining a high inference accuracy}, remains an open question. 
\section{Design Methodologies}\label{sec:design}

In this section, we describe in detail  \gls{deepsz}, our proposed lossy compression framework for neural networks.

\begin{figure*}[ht]
    \vspace{-5mm}
    \centering
    \includegraphics[trim={0 0 0 3mm}, width=0.8\linewidth]{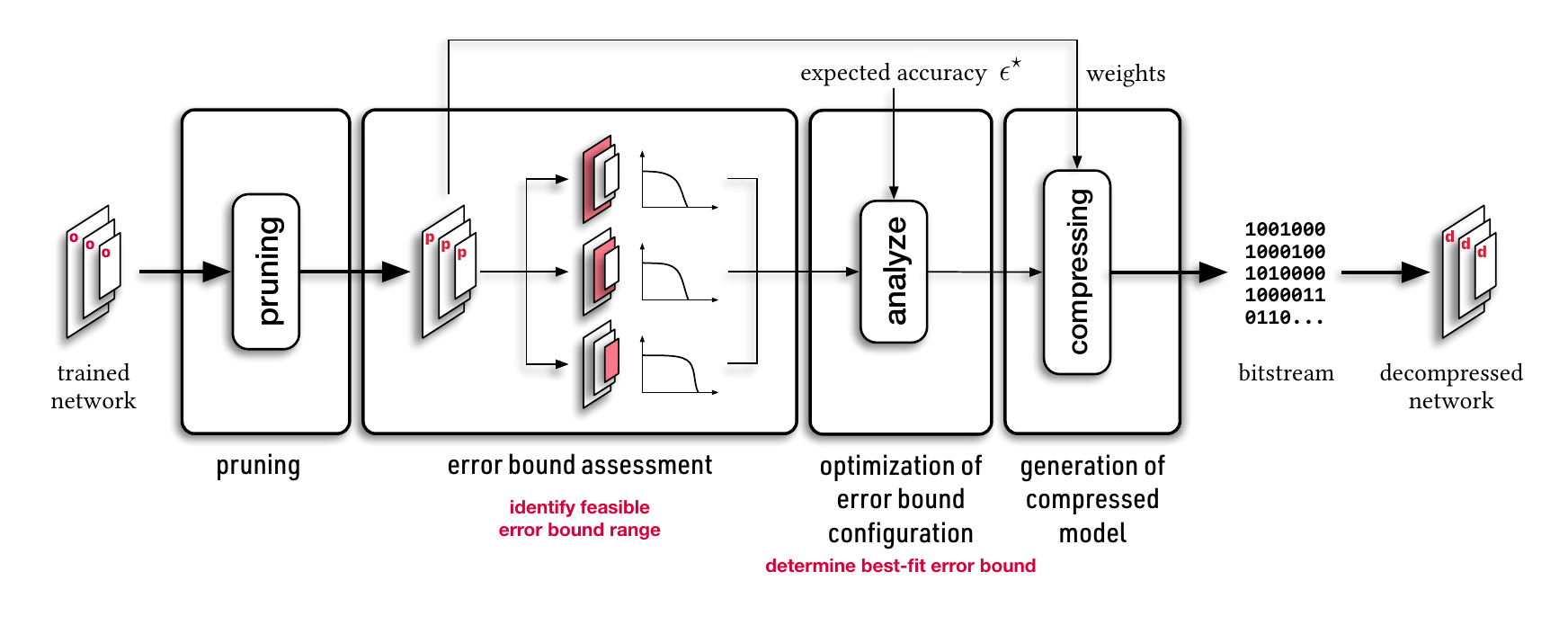}
    \vspace{-5mm}
    \caption{Overview of \textsc{DeepSZ} framework for neural network compression.}
    \label{fig:framework}
\end{figure*}

\subsection{Overview of \textsc{DeepSZ} Framework}\label{sec:design_01_overview}

The general workflow of the \gls{deepsz} framework is presented in Figure~\ref{fig:framework}. As illustrated in the figure, \gls{deepsz} consists of four key steps:  \textit{network pruning}, \textit{error bound assessment}, \textit{optimization of the error bound configuration}, and \textit{generation of the compressed model}. The first step is to adopt  network pruning in order to reduce the network complexity and mitigate the overfitting problem caused by the large number of parameters in the network. The second step is to apply the error-bounded lossy compression to the pruned \glspl{fclayer} and assess the impacts of different error bounds on the inference accuracy for different \glspl{fclayer}. Based on the inference accuracy degradation, \gls{deepsz} will identify the feasible range of error bounds for each \gls{fclayer} and collect the results of inference accuracy degradation and compressed layer size based on these bounds. This step can effectively narrow the range of the best-fit error bounds for each layer. 
Note that we focus only on the \glspl{fclayer} in this work because it dominates most of the storage space, as discussed in Section~\ref{sec:background}.
The third step is to determine the best-fit error bound for each \gls{fclayer} based on the narrowed feasible range generated from the second step. \gls{deepsz} will compress the network as much as possible while satisfying the user-set inference accuracy requirement. 
The fourth step is to generate the compressed network based on the optimized error bounds and best-fit lossless compressor.
In the remainder of this section, we  discuss each step of \gls{deepsz} in detail.

\subsection{Network Pruning}\label{sec:design_02_pruning}
An \gls{fclayer} in DNNs can be represented by a floating-point matrix. Each nonzero element in the matrix represents the weight of one connection between previous layer and current layer. Previous studies on modern neural network models \cite{lecun1990optimal, hassibi1993second} have shown that most of the weights in \glspl{fclayer} are redundant and can be pruned without any impact on the inference accuracy. Moreover, network pruning is an effective way to prevent the DNN model from overfitting. 

We build our pruning method on top of  prior state-of-the-art techniques \cite{han2015deep}, which can prune DNNs without loss of inference accuracy.
We first set up thresholds for each \gls{fclayer} 
and prune their weights based on these thresholds: every weight below these thresholds will be removed.
The thresholds are set based on the predefined pruning ratios suggested by previous studies \cite{han2015learning}.
Then, we retrain the network with masks (i.e., zero weights are marked as unchanged) on \glspl{fclayer} such that the weights that have been pruned can be kept zero. This pruning method is called \emph{magnitude threshold plus retraining} (\textsc{Magnitude}). 
Note that this process can start from well-trained networks;  more details can be found in \cite{han2015learning}.
We note that Reagen et al. \cite{reagen2017weightless} presented another pruning method, dynamic network surgery (\textsc{DNS}), and evaluated its performance on several networks.
The time overhead of DNS applied to large networks (such as VGG-16) is very high, however, because DNS needs to iteratively prune the weights and retrain the network based on increasing thresholds. In contrast, \textsc{Magnitude} has relatively lower time overhead thus can be well applied to large neural networks. Therefore, we focus on the \textsc{Magnitude} method in this paper.


After the network pruning, the weight matrix becomes sparse, so it can be represented by a sparse matrix format, such as \textsc{compressed sparse row} (CSR) or \textsc{compressed sparse column} (CSC) format.
Unlike the traditional format that uses three 1-D arrays (e.g., arrays for nonzero values, the extents of rows, and column indices in CSR), we only use two 1-D arrays to represent one \gls{fclayer} after the pruning.  
One array is named \gls{dataArray} \footnote{We note that the nonzero floating-point weights will be condensed into a 1-D array or linked list regardless of the sparse matrix representation format.};  it is used to store the floating-point weights (32 bits per value); the other one is named \gls{indexArray}; it is used to store the index differences between two consequent nonzero weights (8 bits per value). Similar to \cite{han2015deep}, if the index difference exceeds 256 (i.e., $2^8$), we additionally save a zero padding to \gls{dataArray} and $255$ to \gls{indexArray}.
Here we use sparse matrix representation because the inference accuracy can be dropped sharply (i.e., to 20\% on the tested networks) if the lossy compression is applied to the matrices of pruned weights (i.e., 2-D arrays) based on our experiments. Note that the real compression ratio after the pruning step (i.e., original size divided by the CSR size) is always lower than the compression ratio that we set for the pruning (i.e., one divided by the pruning ratio), because after the pruning every nonzero weight will be represented by 40 bits (8 for index and 32 for data), which is slightly larger than the original 32 bits. Based on our evaluation results, the pruning step can typically reduce the size of \glspl{fclayer} by about 8$\times$ to 20$\times$ if the pruning ratio is set to be around 4$\%$ to 10$\%$.


\subsection{Error Bound Assessment}\label{sec:design_03_assessment}
\label{sec:err-bound-assessment}

\begin{figure}[t]
    \includegraphics[width=0.8\linewidth]{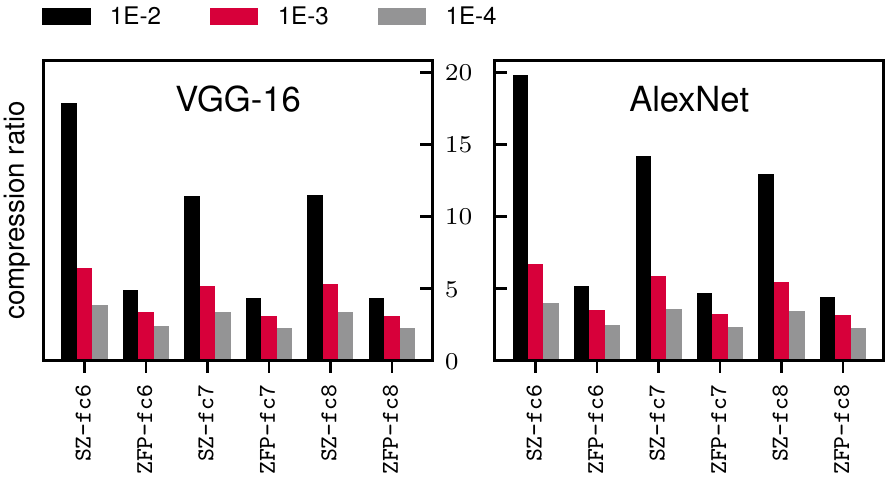}
    \vspace{-2mm}
    \caption{Compression ratios of SZ and ZFP lossy compression on \gls{fclayer} in AlexNet and VGG-16.}
    \label{fig:CR_SZ_ZFP}
\end{figure}

\gls{szfull} usually has a much higher compression ratio on 1-D datasets than other lossy compression methods do \cite{di2016fast, tao2017depth}. Our floating-point datasets are 1-D \glspl{dataArray}, as described in Section~\ref{sec:design_02_pruning}.
For demonstration purposes, we evaluated SZ and ZFP\footnote{Many recent studies, such as \cite{tao2018optimizing}, have demonstrated that SZ and ZFP are two leading lossy compressors for floating-point data.} on the 1-D \glspl{dataArray} of each \gls{fclayer} in AlexNet and VGG-16. The compression ratios are presented in Figure~\ref{fig:CR_SZ_ZFP}. The figure shows that SZ consistently outperforms ZFP in terms of compression ratios on the tested \glspl{fclayer} with  absolute error bounds of $10^{-2}$, $10^{-3}$, and $10^{-4}$. 
Although SZ has higher compression and decompression times than does ZFP \cite{tao2017significantly}, they are still much lower than the time overheads of forward or backward pass in neural networks. 
Taking these facts into consideration, we propose to use SZ lossy compression in our \gls{deepsz} framework. 

\begin{figure}[t]
  \includegraphics[trim={0 4mm 0 3mm}, clip, width=\linewidth]{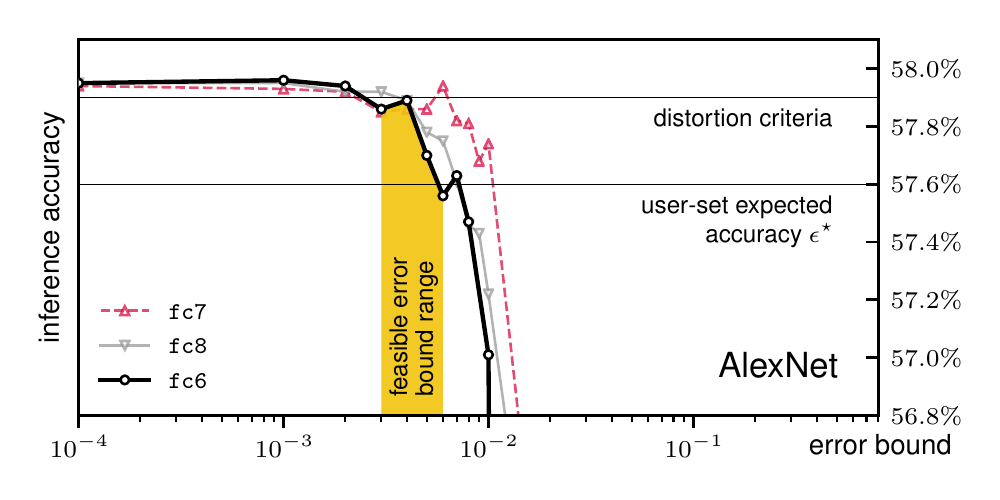}
  \caption{Inference accuracy of different error bounds on the \glspl{fclayer} in AlexNet.}
  \label{fig:feasible-range}
\end{figure}

Error-bounded lossy compression can provide high compression ratio but can also bring bounded errors to neural networks, leading to possible loss of inference accuracy. 
Thus, before adopting SZ to compress \glspl{fclayer}, we need to find the best-fit error bound for each layer. Our idea is to narrow the best-fit error bounds by identifying a feasible error bound range for each \gls{fclayer} with a high compression ratio and also bounded loss of inference accuracy (detailed in this subsection) and then fine tune the best-fit error bound within the range (detailed in next subsection).
To this end, we need to understand the impact of different error bounds of each \gls{fclayer} on the overall inference accuracy. Specifically, we use SZ to compress each \gls{fclayer}'s \gls{dataArray} with different error bounds and use its decompressed array to reconstruct the \gls{fclayer} while leaving the other \glspl{fclayer} uncompressed or unchanged. Based on the reconstructed network (only one \gls{fclayer} is modified), we can perform the forward pass on the test data to generate the inference accuracy. Thus, we can get a series of inference accuracies based on different error bounds for each \gls{fclayer} in the network. For example, Figure~\ref{fig:feasible-range} presents the accuracies based on the absolute error bounds from $10^{-1}$ to $10^{-4}$ for the three \glspl{fclayer} in AlexNet. Note that in order to reduce the time overhead, we choose only a group of error bounds for compression, decompression, and checking inference accuracy rather than all the error bounds shown in Figure \ref{fig:feasible-range}. In the following text we will describe how to determine the error bounds.



In our solution, we test the inference accuracy with only one compressed layer in every test, instead of using a brute-force method to search all possible test cases (i.e., all combinations of error bounds across all layers), for the following two reasons. (1) We observe that the \glspl{fclayer} in neural networks usually have \textit{independent} characteristics in the context of SZ lossy compression. That is, two reconstructed \glspl{fclayer} based on SZ affect the overall accuracy independently.
Thus, the overall loss of inference accuracy can be composed of (and thus estimated by) the losses of inference accuracy introduced by individual layers. 
We will discuss more details in Section~\ref{sec:design_04_optimize}.
(2) Checking the inference accuracy in a test with only one reconstructed layer using multiple error bounds has much less computational cost than does a brute-force method involving every possible combination of error bounds across multiple layers. Specifically, our solution has a linear time complexity compared with the brute-force method with an exponential time complexity. VGG-16, for instance, has three \glspl{fclayer}. Assume we have 10 candidate error bounds for each layer. Then  the brute-force method needs to check 1,000 test cases, each involving one compression, one decompression, and one forward-pass test. By comparison, our solution has only 30 test cases to check, thus reducing the testing time to 3\% compared with the brute-force method.

\begin{algorithm}[t]
    \footnotesize
    \caption{Error bound assessment for \glspl{fclayer} in deep neural networks.}
    \label{algo:fc-layer}
    \parbox[l]{\linewidth}{\hspace*{\algorithmicindent}%
    {\bfseries Notation:} %
        layer: $\ell$; \textsc{Accuracy Degradation}: $\Delta$; error bound: $eb$; threshold: $\tau$; Expected Accuracy Loss: $\epsilon^\star$; the given network: $\mathcal{N}$; size of compressed layer $\ell$ with $eb$: $\sigma_{(\ell; eb)}$
    }
\parbox[l]{\linewidth}{\hspace*{\algorithmicindent}%
    {\bfseries Global:} %
        $\mathcal{N}$, $\epsilon^\star$, all $\sigma_{(\ell;eb)}$, all $\Delta_{(\ell; eb)}$
    }
\begin{algorithmic}[1]
\Procedure{Check}{$\ell^\circ$, $eb^\circ$, base}
    \State test $\Delta_{(\ell^\circ; eb^\circ)}$ on $\mathcal{N}$ and update $\sigma_{(\ell^\circ;eb^\circ)}$
    \If{%
        $\Delta_{(\ell^\circ; eb^\circ)}>\epsilon^\star$%
    } 
        \State terminate the $\textsc{Check}$
    \Else
        \State $eb^\circ \gets eb^\circ+$base 
        \State base $\gets 10 \times\text{base}$ \textbf{if} $eb^\circ = 10\times\text{base}$
        \State \textsc{Check}($\ell^\circ$, $eb^\circ$, base)
    \EndIf
\EndProcedure
\State

\Procedure{ErrorBoundAssessment}{}
    \For{$\ell \gets$ layers in network}
        \For{$\beta \gets$ {\small[1E-3,1E-2,1E-1]}} \Comment{can be pushed to 1E-4 etc.}
        \If{$\Delta_{(\ell; \beta)} > 0.1\%$}
            \State \textsc{Check}($\ell$, $1\times\beta/10$, $\beta/10$)
            \State break
        \EndIf
        \EndFor
        \State \Return all $\sigma_{(\ell;eb)}$, all $\Delta_{(\ell; eb)}$
    \EndFor
\EndProcedure
\end{algorithmic}
\end{algorithm}

We propose an algorithm to identify the feasible range of error bounds and collect the results about inference accuracy degradation and compressed layer size based on these bounds for each \gls{fclayer}.
We present the pseudo-code in Algorithm~\ref{algo:fc-layer}. The inputs of the algorithm include the architecture of the neural network, the pruned weights of the network, and the user-set loss of inference accuracy. Lines \textcolor{RED}{\ttfamily 12}--\textcolor{RED}{\ttfamily 21} show the main loop of this algorithm. Specifically, when the loss of inference accuracy exceeds a criterion of 0.1\% (called distortion criterion) in terms of absolute percentage, we treat the reconstructed network to be distorted.
We search the feasible range of error bounds by checking the accuracy based on multiple error bounds. The error bound to check starts from a certain value (i.e., the default value is $10^{-3}$; the reason will be discussed in Section 5.1) and will be increased by an order of magnitude (10$\times$) each time. As the accuracy drops below the criterion (i.e., 0.1\%) at a certain error bound $eb^\prime$, we set the starting point of the range to be $eb^\prime/10$. Note that the default value of $10^{-3}$ can be further decreased, such as $10^{-4}$, based on different neural networks.

From Lines \textcolor{RED}{\ttfamily 1}--\textcolor{RED}{\ttfamily 10}, we determine the ending point of the range based on the accuracy of reconstructed network. The ending point is the first error bound that the accuracy drops below the user's expected accuracy $\epsilon^*$.
Once the feasible range is generated (as shown in Figure~\ref{fig:feasible-range}), we conduct the tests with the error bounds in the feasible range and collect the sizes of compressed layer and accuracy degradation results. Lines \textcolor{RED}{\ttfamily 6}--\textcolor{RED}{\ttfamily 8} describe how we choose the error bounds within the range. For example, if the range is $[8\times10^{-3}$, $3\times10^{-2}$], we test the error bounds of $8\times10^{-3}$, $9 \times 10^{-3}$, $10^{-2}$, $2\times10^{-3}$, and $3\times10^{-3}$. 



\subsection{Optimization of Error Bound Configuration}\label{sec:design_04_optimize}
\label{sec:fine-turning}

Depending on the number of input and output neurons, the sizes of different \glspl{fclayer} vary dramatically.
For example, the largest \gls{fclayer} of VGG-16 is \texttt{fc6} (i.e., 25,088$\times$4,096), which is 25$\times$ larger than the smallest layer \texttt{fc8} (i.e., 1,000$\times$4,096). 
Compressing larger \glspl{fclayer} with a higher error bound can lead to a higher compression ratio, which can benefit the overall ratio. However, the higher error bound also brings more errors to the network, which may degrade the overall accuracy in turn. Therefore, how to determine the optimal error bound for each \gls{fclayer} is an important problem.

From our experiments we discovered that the overall accuracy loss in the neural network exhibits an approximate linearity in terms of the accuracy degraded in each \gls{fclayer} in the context of SZ lossy compression when the targeted loss of inference accuracy is lower than 2\% based on our tested neural networks. 
In other words, the overall accuracy loss $\Lambda^\star$ is approximately equal to the sum of each layer's accuracy degradation $\Delta_\ell$, as shown in Equation~(\ref{eq:ADsum}), where $eb_\ell$ is a given arbitrary error bound for each \gls{fclayer} $\ell$:
\begin{equation}\label{eq:ADsum}
    \Delta^\star = \sum \Delta_{\ell, eb_\ell},  \qquad   \Delta^\star<2\%.
\end{equation}

This observation can be explained by a theoretical analysis on the independent impact of compression error introduced in each \gls{fclayer} on the output of a neural network. Assume $t_1$ and $t_2$ to be the original outputs of two successive \glspl{fclayer}. Assume $\Delta W_1$ and $\Delta W_2$ to be the compression errors introduced to the weights of these two \glspl{fclayer}, respetively. When computations pass these two \glspl{fclayer}, the outputs of the two \gls{fclayer} would be
\begin{equation}\label{eq:actFun}
    t_1^\star = f((W_1 + \Delta W_1)t_0 + b_1), \quad t_2^\star = f((W_2 + \Delta W_2)t_1^\star + b_2),
\end{equation}
where $f$ represents the activation function of \glspl{fclayer}, $W$ represents the weights, and $b$ represents the bias. For most of neural networks, the activation function $f$ used in \glspl{fclayer} is rectified linear unit (a.k.a., ReLU), which is $max(0, x)$ (where $x$ is the input to a neuron). In this case, due to $\Delta W_1 << W_1$ and $\Delta W_2 << W_2$, Equation (\ref{eq:actFun}) can be simplified to
\begin{align}
    t_1^\star &= f(W_1t_0 + b_1) + f\Delta W_1t_0 = t_1 + f\Delta W_1t_0, \\
    t_2^\star &= t_2 + fW_2\Delta W_1t_0 + f\Delta W_2t_1.    
\end{align}
The final output $T^\star$ of the network would be 
\begin{equation}
    T^\star = T + f\Delta W_1t_0\Pi^{n}_{i=2}(W_i) + f\Delta W_2t_1\Pi^{n}_{i=3}(W_i).
\end{equation}
where $n$ is the total number of \glspl{fclayer} and $T$ is the original output. We can clearly observe that two errors $\Delta W_1$ and $\Delta W_2$ have independent impacts on the final output $T^\star$. Therefore, we can conclude that compression error introduced in each \gls{fclayer} would have independent impact on final network's output. In order to assure $\Delta W << W$, we will choose the error bound to be lower than $0.1$ in our experiments. Finally, the relationship between final output and accuracy loss is approximately linear, which will be experimentally demonstrated in Section \ref{sec:evaluation-results}.


We propose Algorithm 2 to determine the best-fit error bound for each layer. The inputs include the accuracy degradation and compressed size of each \gls{fclayer} based on our tested error bounds, which are outputted by our previous error bound assessment (Section \ref{sec:err-bound-assessment}) and an expected loss of accuracy set by users. Algorithm 2 can minimize the total size of the compressed \glspl{fclayer} while ensuring the sum of the accuracy degradation of each layer to be within the expected accuracy loss.
Specifically, the first part of Algorithm 2 (Lines \textcolor{RED}{\ttfamily 2}--\textcolor{RED}{\ttfamily 14}) finds the minimum total size of compressed \glspl{fclayer} with different combinations of error bounds before a certain layer by using a variation of Knapsack algorithm. 
Then, the algorithm traces back to determine the error bound for each \gls{fclayer} (Lines \textcolor{RED}{\ttfamily 15}--\textcolor{RED}{\ttfamily 19}). More specifically, we save the minimal size of all \glspl{fclayer} (before $\ell$-layer) and the accuracy loss of $\epsilon^*$ to the variable $S_{(\ell;\epsilon)}$. After we find the minimum compressed size of all the \glspl{fclayer} under the constraint of the overall accuracy loss (Lines \textcolor{RED}{\ttfamily 13}--\textcolor{RED}{\ttfamily 14}), we trace back from the minimal $S_{(\ell;\epsilon)}$ to identify the error bound combination for each layer (Lines \textcolor{RED}{\ttfamily 15}--\textcolor{RED}{\ttfamily 19}).

\begin{algorithm}[t]
    \footnotesize
    \caption{Optimization of Error Bound Configuration}
    \label{algo:program}
    \newcommand{\AccuracyDegradation}{{\sc AccuracyDegradation}}
\newcommand{\ExpectAccuracy}{{\sc ExpectAccuracy}}
\newcommand{\ErrorBound}{{\sc ErrorBound}}
\newcommand{\Size}{{\sc Size}}
\newcommand{\TotalSize}{{\sc TotalSize}}

\parbox[l]{\linewidth}{\hspace*{\algorithmicindent}%
    \textbf{Notation:} layer: $\ell$; error bound: $eb$; accuracy: $\epsilon$; \AccuracyDegradation\ at $\ell$ with $eb$: $\Delta_{(\ell; eb)}$; size: $\sigma$; \TotalSize: $S$; \ExpectAccuracy: $\epsilon^\star$
    }
\parbox[l]{\linewidth}{\hspace*{\algorithmicindent}%
    \textbf{Input:} $\Delta_{(\ell;eb)}$, %
    $\epsilon^\star$, %
    $\sigma_{(\ell; eb)}$ %
    }
\makebox[\linewidth][l]{\hspace*{\algorithmicindent}%
    \textbf{Output:} $eb_{\ell}$
    }
\begin{algorithmic}[1]
\Procedure{OptimizeErrorBound}{}
    \State $S_{(\ell; \Delta)} \gets$ maximum
    \State $S_{(\ell_\text{zero}; \epsilon)} \gets$ zero
    \For{$\ell \gets$ layers in network}
        \For{$eb \gets$ tested error bounds}
            \For{$\epsilon \gets [0\ldots 100]\times\epsilon^\star$}
                \If{$S_{\left(\ell_\text{prev}; \epsilon\right)}+ \sigma_{(\ell; eb)} < S_{\left(\ell; \epsilon+\Delta_{(\ell,eb)}\right)}$ }
                    \State $S_{\left(\ell; \epsilon+\Delta_{(\ell, eb)}\right)} \gets S_{(\ell_\text{prev}; \epsilon)}+ \sigma_{(\ell; eb)}$
                \EndIf
            \EndFor
        \EndFor
    \EndFor
    \State let $\ell^\star$ be the final layer in the network
    \State find minimum $S_{(\ell^\star; \Delta)}$
    \For{$\ell \gets$ layers in network}    \Comment{tracing back}
        \For{$eb \gets$ tested error bound}
            \State $eb_\ell \gets eb$ \textbf{if} $S_{(\ell; \epsilon)}- \sigma_{(\ell; eb)} = S_{\left(\ell_\text{prev}; \epsilon-\Delta_{(\ell; eb)} \right)}$
        \EndFor
    \EndFor
\EndProcedure
\end{algorithmic}
\end{algorithm}

Besides this optimization of the compression ratio by an expected accuracy loss (i.e., expected-accuracy mode), \gls{deepsz} can optimize the overall accuracy with an expected compression ratio (i.e., expected-ratio mode).
The algorithm of the fixed-rate mode is similar to Algorithm 2 but just reverses the compressed size and accuracy degradation. Based on these two  modes, we can fine tune the balance between accuracy loss and compression ratio for a neural network, which is much more flexible than other methods.


\subsection{Generation of Compressed Model}\label{sec:design_05_generation}

\begin{figure}
    \includegraphics[width=0.8\linewidth]{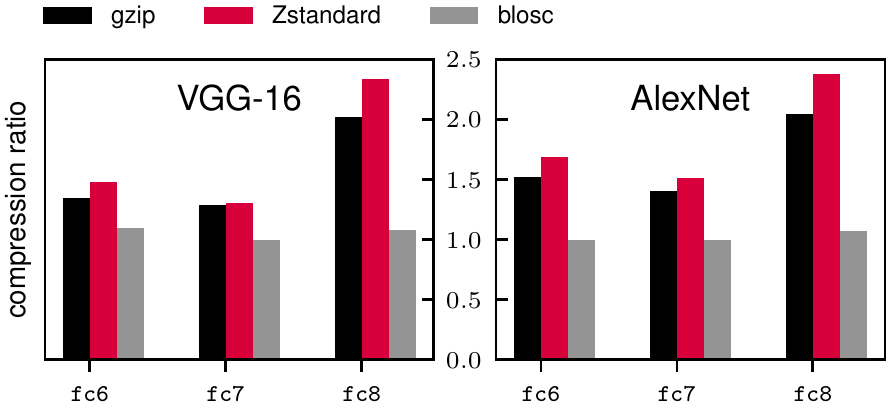}
    \caption{Compression ratios of different layers' \glspl{indexArray} with different lossless compressors on AlexNet and VGG-16.}\label{fig:CompressCompare}
\end{figure}

The last step in our framework is to generate the compressed model by using SZ lossy compression on the \glspl{dataArray} with the error bounds (obtained in Step 3) and the best-fit lossless compression on the \glspl{indexArray}.
The \gls{indexArray} represents the locations of nonzero weights, which need to be compressed losslessly.
\gls{deepsz} provides three state-of-the-art lossless compressors: Gzip \cite{deutsch1996gzip}, Zstandard \cite{zstd}, and Blosc \cite{zstd}. More lossless compressors can be integrated into the framework in the future. 
In our experiments, we identified that Zstandard always leads to the highest compression ratio compared with the other two compressors, as shown in Figure~\ref{fig:CompressCompare}. After these four steps of \gls{deepsz}, the compressed neural network model is generated. In this paper, we use \textsc{encoding} to refer this whole process of generating compressed DNNs and \textsc{decoding} to refer the process of reconstructing DNNs.

Once the network is needed for forward pass, it must be decoded. During the decoding, \gls{deepsz} will decompress the \glspl{dataArray} using the SZ lossy compression and the \glspl{indexArray} using the best-fit lossless compression (e.g., Zstandard). Then, the sparse matrix can be reconstructed based on the decompressed \gls{dataArray} and \gls{indexArray} for each \gls{fclayer}. Finally, the whole neural network can be decoded.
Note that the computational cost of the decoding in \gls{deepsz} is relatively low compared to that of the forward pass with a batch of images.
We will analyze the performance overhead of our decoding in detail and compare it with other state-of-the-art methods next section.

\section{Detailed Analysis and Comparison}
\label{sec:comparison}
In this section, we analyze \gls{deepsz} in detail and compare it with two other state-of-the-art solutions:  Weightless \cite{reagen2017weightless} and Deep Compression \cite{han2015deep}. Our analyses focus on both performance and storage.


\subsection{Performance Analysis of \textsc{DeepSZ}}
For Algorithm 1 in \gls{deepsz}, the computational cost is focused mostly on performing the tests  with different error bounds to check the corresponding accuracies, while compression and decompression both cost negligible time overhead.
Let us take AlexNet as an example. Compressing and decompressing one \gls{dataArray} (about tens of megabytes per \gls{fclayer}'s \gls{dataArray}, as shown in Table~\ref{tab:subtable_AlexNet}), and reconstructing the network based on the decompressed layer typically take no more than one second on an Nvidia Tesla V100 GPU,\footnote{Based on a recent study \cite{tao2017significantly}, SZ's compression and decompression rate are about 80 MB/sec and 150 MB/sec, respectively, with the error bound of $10^{-3}$ on a 2.3 GHz Intel Core i7 processor.} whereas testing the reconstructed network with 50,000 images in the ImageNet dataset will take about 55 seconds (10 seconds for data transfer, 5 seconds for initialization, and 40 seconds for forward computations).
In this case, \gls{deepsz} needs to perform 12 tests on each \gls{fclayer}, - bringing the total to 36 tests. In contrast, the 36 tests require performing forward passes of 1.8 million images considering 50,000 images in the ImageNet test data. Based on our experiments, the execution time of one epoch is about 42 times higher than that of one \textsc{test} with an Nvidia V100 GPU on AlexNet \footnote{One epoch contains one forward pass and one backward pass of the 1.28 million images in the training dataset and takes about 15.7 minutes for AlexNet on a single Nvidia Tesla V100 GPU, based on our experiments.}. Thus, the workload of 1.8 million images in the \textsc{test} is equivalent to training about $\frac{6}{7}$ epochs of data (i.e., one \textsc{test} is about $\frac{1}{42}$ epochs). Therefore, the time complexity of Algorithm 1 is $\mathcal{O}\big(\tfrac{c\cdot k \cdot M}{42}\big)$, where $c$ is the number of tests per layer (e.g., 12 for AlexNet), $k$ is the number of \glspl{fclayer} (e.g., 3 for AlexNet), and $\mathcal{O}(M)$ is the time complexity for training one epoch of data. For AlexNet, for example, we can set $c$ to 12 and $k$ to 3; hence, the time complexity is about $\mathcal{O}(\frac{6\cdot M}{7})$.


For Algorithm 2 in \gls{deepsz}, because of our optimization in Algorithm 1, the input dimension of the algorithm is very small (i.e., $\mathcal{O}(c\cdot k)$, for example 36 pairs of inference accuracy degradation and compressed size for 3-\gls{fclayer} AlexNet. Based on the time complexity of the Knapsack algorithm, the time complexity of Algorithm 2 is $\mathcal{O}(100\!\cdot\!\epsilon^\star\!\! \cdot\! c\! \cdot\! k )$. We note that $\mathcal{O}(100\!\cdot\!\epsilon^\star\!\! \cdot\! c\! \cdot\! k )$ is far smaller than $\mathcal{O}\big(\tfrac{c\cdot k \cdot M}{42}\big)$ because $\mathcal{O}\big(\tfrac{M}{42}\big)$ is much larger than $\mathcal{O}(100\cdot\epsilon)$. 
Thus, the computational cost of Algorithm 2 would be relatively small compared with multiple tests of inference accuracy in Step 2. Overall, we conclude that the time complexity of \gls{deepsz}'s encoding is $\mathcal{O}\big(\tfrac{c\cdot k \cdot M}{42}\big)$, which is much less than  that of traditional methods with retraining (typically $\mathcal{O}(5\cdot M)$ to $\mathcal{O}(10\cdot M)$ in the ImageNet dataset).
It is worth noting that the scalability of \textsc{test} (i.e., embarrassing parallelism) is higher than that of training in parallel, thus, the time complexity of \gls{deepsz} compared with training will be further reduced with increasing scale. 



For \gls{deepsz}'s decoding, the computational cost is also comparatively low because it performs an $\mathcal{O}(n_\text{pruned})$ lossy decompression with SZ, an $\mathcal{O}(n_\text{pruned})$ lossless decompression with the best-fit lossless compressor (e.g., Zstandard) and an $\mathcal{O}(n)$ sparse-dense matrix conversion. Here we denote the number of pruned weights by $n_\text{pruned}$ and the number of original weights by $n$. Overall, the time complexity of \gls{deepsz}'s decoding is $\mathcal{O}(n)$. 

\subsection{Comparison with Weightless}


\gls{deepsz} has four major advantages over Weightless. (1) Weightless has higher time overhead than does \gls{deepsz} for encoding. After Weightless reconstructs the layer based on the Bloomier filter, the inference accuracy can drop dramatically. For example, the inference accuracy drops about 3\% when compressing \texttt{fc6} in VGG-16 using Weightless. Thus, Weightless requires  retraining the other layers to recover the overall inference accuracy, whereas \gls{deepsz} does not require any retraining.
(2) Weightless has higher time overhead than does \gls{deepsz} on decoding. To decode one element, Weightless has to calculate four hash functions based on all the values (including zero values) in the pruned matrix and check the hash table to determine the value of this element, leading to much higher time overhead compared with \gls{deepsz}.
(3) Weightless  can  compress only one layer (usually the largest layer). By contrast, \gls{deepsz} can compress all \glspl{fclayer}, leading to higher overall compression ratio. 
(4)  \gls{deepsz}  provides two  modes to users. Even for the fixed-accuracy mode, users can set an expected loss of inference accuracy in \gls{deepsz} and get as high a compression ratio as possible, whereas Weightless is unable to provide such flexibility.


\subsection{Comparison with Deep Compression}

Similar to Weightless, Deep Compression also requires retraining the whole network to mitigate the inference accuracy loss caused by its quantization.
Deep Compression adopts a simple quantization technique on the pruned weights. It quantizes all the nonzero weights to a group of floating-point values based on a code-book. The number of these values in the code-book is always $2^k$, where $k$ refers to the number of bits used to represent one weight. Using 5 bits per weight, for example, can map every nonzero weights to a 32-value code-book.
Unlike Deep Compression applying a simple quantization to the weights, \gls{deepsz} applies an error-bounded linear-scaling quantization to the difference between the predicted weight and real weight based on a best-fit prediction method, leading to higher compression ratios and fine-granularity error controls.
Similar to Weightless, Deep Compression has lower flexibility than \gls{deepsz} has in terms of the balance between the ratio and inference accuracy. Since the number of floating-point values the code-book can represent is always $2^k$,  
the inference accuracy under Deep Compression may drop significantly (shown in Section 5.2) with increasing compression ratios (or lower bit rates), leading to unbounded inference accuracy.


\section{Experimental Evaluation}
\label{sec:evaluation}

In this section, we evaluate our proposed \gls{deepsz} framework by comparing it with  state-of-the-art methods.

\begin{figure}[]
	\begin{subfigure}{\linewidth}\centering
	    \includegraphics[width=0.87\linewidth]{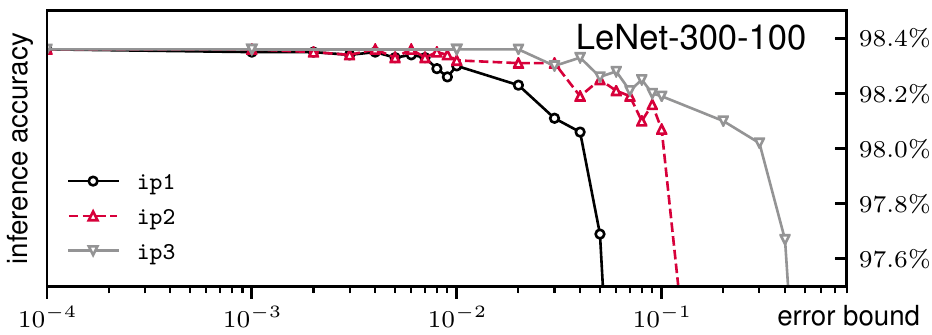}
	    \caption{\footnotesize LeNet-300-100}\label{fig:subfigure_AccuracyDrop_LeNet-300-100}
	    \vspace{4mm}
	\end{subfigure}
	\begin{subfigure}{\linewidth}\centering
	    \includegraphics[width=0.87\linewidth]{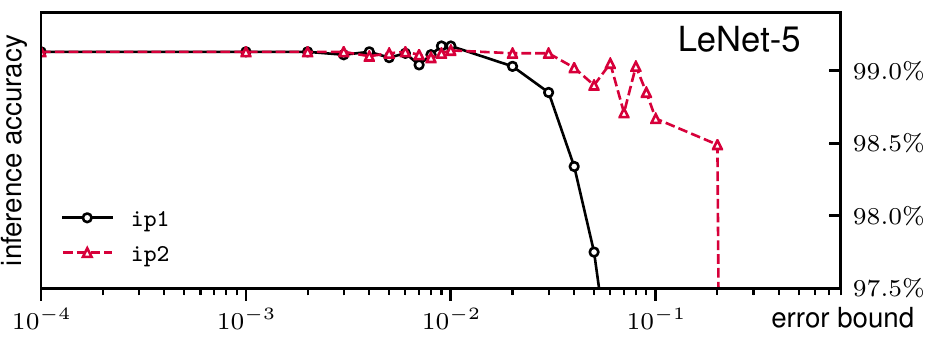}
	    \caption{\footnotesize LeNet-5}\label{fig:subfigure_AccuracyDrop_LeNet-5}
	    \vspace{4mm}
	\end{subfigure}
	\begin{subfigure}{\linewidth}\centering
	    \includegraphics[width=0.87\linewidth]{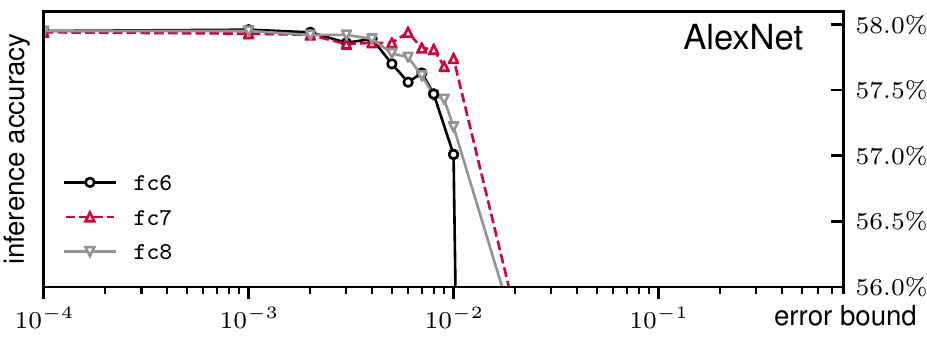}
	    \caption{\footnotesize AlexNet}\label{fig:subfigure_AccuracyDrop_AlexNet}
	    \vspace{4mm}
	\end{subfigure}
	\begin{subfigure}{\linewidth}\centering
	    \includegraphics[width=0.87\linewidth]{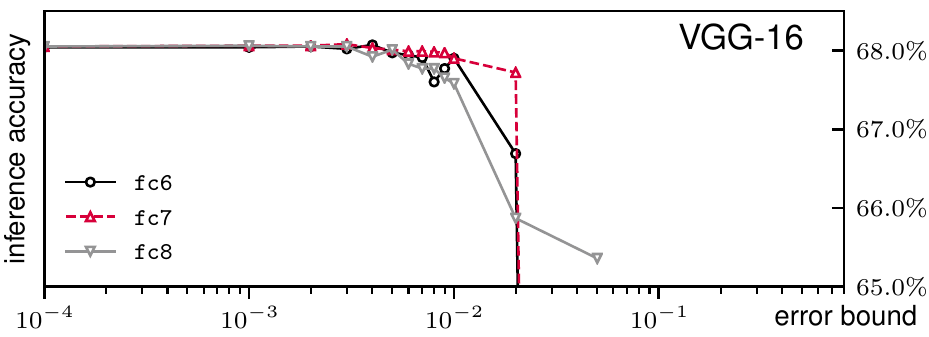}
	\caption{\footnotesize VGG-16}\label{fig:subfigure_AccuracyDrop_VGG-16}
	\end{subfigure}
\caption{Inference accuracy of different error bounds on the \glspl{fclayer} in LeNet300-100, LeNet-5, AlexNet, and VGG-16}
\label{fig:error-bound-assessment}
\end{figure}

\subsection{Experimental Setting}

We conduct our evaluation on a single core of an MacBook Pro with Intel Core i7-8750H Processors (with 32 GB of memory) and parallel experiments using four Nvidia Tesla V100 GPUs (each with 16 GB of memory) on the node of the Pantarhei cluster at the University of Alabama. The four GPUs are connected via NVLink \cite{foley2017ultra}. 
We implement  \gls{deepsz} based on the Caffe deep learning framework \cite{Jia:2014up} (v1.0) and SZ lossy compression library (v2.0) \cite{liangerror}.
We evaluate \gls{deepsz} on four well-known neural networks: LeNet-300-100 \cite{lecun1998gradient}, LeNet-5 \cite{lecun2015lenet}, AlexNet \cite{krizhevsky2012imagenet}, and VGG-16 \cite{simonyan2014very}.
We train/test LeNet300-100 and LeNet-5 on the MNIST dataset and AlexNet and VGG-16 on the ImageNet dataset \footnote{ImageNet Large Scale Visual Recognition Challenge 2012 (ILSVRC2012)}, respectively.
These neural networks and datasets are commonly used in evaluation studies \cite{han2015deep, reagen2017weightless}.
We present the details of their architectures in Table~\ref{tab:NeuralNetworks}. Note that the \glspl{fclayer} occupy most of the storage space (i.e., 89.4\% $\sim$ 96.1\%).
We use the default solver (i.e., stochastic gradient descent (SGD)) in Caffe for all training.
We set the expected loss of inference accuracy to 0.2\% for two LeNets and 0.4\% for AlexNet and VGG-16, without loss of generality.
We also set the expected loss of inference accuracy to zero and demonstrate the flexibility of \gls{deepsz}.

We note that in an \gls{fclayer} of a neural network, weights are floating-point numbers between -1.0 and 1.0; more generally, for a trained network, such as AlexNet and VGG-16, the value ranges of their weights are typically between -0.3 and +0.3. Thus, the absolute error bounds in the order of $10^{-1}$ are relatively large compared with the weight values. Consequently, using the error bounds in the order of $10^{-1}$ would significantly affect the overall inference accuracy (i.e., dropped to less than 20\%), as illustrated in Figure~\ref{fig:error-bound-assessment}. We also note that the absolute error bound of $10^{-4}$ can maintain the inference accuracy without any loss for these networks. Thus, we set $10^{-3}$ to be the default value for initial point of the error bound to be checked. It is worth pointing out that the output of a neural network for image recognition is a vector of probabilities for different types. Based on these probabilities, a neural network can predict the type of an image. Thus, the inference accuracy in the context of image recognition and neural network is the precision ratio. Similar to previous studies \cite{han2015deep, reagen2017weightless, tung2018clip}, no recall ratio and F1 score would be measured in our experiments.

\subsection{Evaluation Results}
\label{sec:evaluation-results}

\begin{figure}\centering
	    \includegraphics[trim={0 5mm 0 4mm}, width=0.60\linewidth]{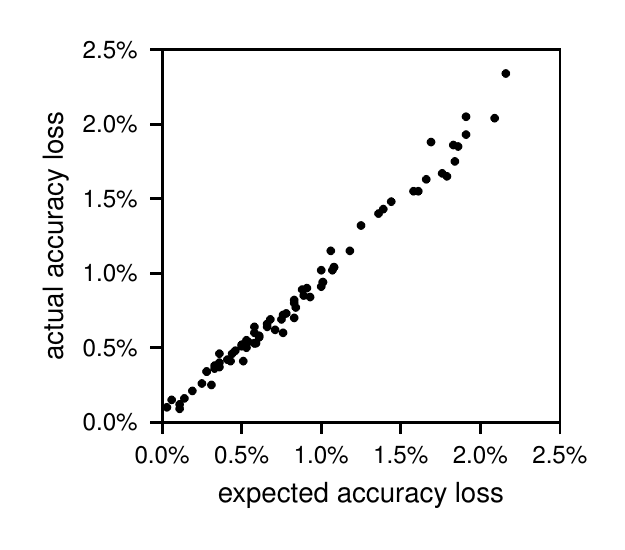}
	    \vspace{-1mm}
	    \caption{Approximate linearity relationships of error introduced by different \glspl{fclayer}}
	    \label{fig:linearity}
\end{figure}

\begin{table}
\caption{\glspl{fclayer}' compression statistics for 4 NNs}
        \begin{subtable}{\linewidth}
        \caption{\glspl{fclayer}' compressing statistics for LeNet-300-100}
        \label{tab:subtable_LeNet-300-100}
        \small\resizebox{\columnwidth}{!}{%
\begin{tabular}{@{}rrrr@{}l@{}r@{}l@{}}
\toprule
\bf Layer  & 
    \bf Original Size  & 
        \bf \begin{tabular}{@{}r@{}}Pruning \\Ratio\end{tabular} & 
            \bf CSR Size  &&
                \bf \begin{tabular}{@{}r@{}}\gls{deepsz} \\Compressed \end{tabular}& \\
\midrule
\texttt{\bfseries ip1} & 941 KB  & 8\%    & 94 KB  & & 15.2 KB  & \\
\texttt{\bfseries ip2} & 120 KB  & 9\%    & 14 KB  & & 1.6 KB & \\
\texttt{\bfseries ip3} & 4 KB    & 26\%   & 1.3 KB & & 0.7 KB & \\
\bfseries overall & 
    1056 KB & 
        8.25\% & 
            109 KB & 
                \ \scriptsize(9.7 $\times$) & 
                    19.1 KB & 
                        \ \scriptsize(55.8 $\times$) \\
\bottomrule
\end{tabular}
}
        \vspace{2mm}
    \end{subtable}
    \begin{subtable}{\linewidth}
        \caption{\glspl{fclayer}' compressing statistics for LeNet-5}
        \label{tab:subtable_LeNet-5}
        \small\resizebox{\columnwidth}{!}{%
\begin{tabular}{@{}rrrr@{}l@{}r@{}l@{}}
\toprule
\bf Layer  & 
    \bf Original Size  & 
        \bf \begin{tabular}{@{}r@{}}Pruning \\Ratio\end{tabular} & 
            \bf CSR Size  &&
                \bf \begin{tabular}{@{}r@{}}\gls{deepsz} \\Compressed \end{tabular}& \\
\midrule
\texttt{\bfseries ip1} & 1600 KB & 8\% & 160 KB && 27.3 KB & \\
\texttt{\bfseries ip2} & 20 KB & 19\% & 4.8 KB && 0.93 KB & \\
\bfseries overall & 1620 KB & 8.1\% & 165 KB & \ \scriptsize(9.8 $\times$) & 28.27 KB & \ \scriptsize(57.3 $\times$) \\
\bottomrule
\end{tabular}
}
        \vspace{2mm}
    \end{subtable}
    \begin{subtable}{\linewidth}
        \caption{\glspl{fclayer}' compressing statistics for AlexNet}
        \label{tab:subtable_AlexNet}
        \small\resizebox{\columnwidth}{!}{%
\begin{tabular}{@{}rrrr@{}l@{}r@{}l@{}}
\toprule
\bf Layer  & 
    \bf Original Size  & 
        \bf \begin{tabular}{@{}r@{}}Pruning \\Ratio\end{tabular} & 
            \bf CSR size  &&
                \bf \begin{tabular}{@{}r@{}}\gls{deepsz} \\Compressed \end{tabular}& \\
\midrule
\texttt{\bfseries fc6} & 151.0 MB & 9\% & 17.0 MB && 2.77 MB & \\
\texttt{\bfseries fc7} & 67.1 MB & 9\% & 7.5 MB && 1.44 MB & \\
\texttt{\bfseries fc8} & 16.4 MB & 25\% & 5.1 MB && 0.94 MB & \\
\bfseries overall & 234.5 MB & 10.1\% & 29.6 MB & \ \scriptsize(7.9 $\times$) & 5.15 MB & \ \scriptsize(45.5 $\times$) \\
\bottomrule
\end{tabular}
}
        \vspace{2mm}
    \end{subtable}
    \begin{subtable}{\linewidth}
        \caption{\glspl{fclayer}' compressing statistics for VGG-16}
        \label{tab:subtable_VGG-16}
        \small\resizebox{\columnwidth}{!}{%
\begin{tabular}{@{}rrrr@{}l@{}r@{}l@{}}
\toprule
\bf Layer  & 
    \bf Original Size  & 
        \bf \begin{tabular}{@{}r@{}}Pruning \\Ratio\end{tabular} & 
            \bf CSR Size  &&
                \bf \begin{tabular}{@{}r@{}}\gls{deepsz} \\Compressed \end{tabular}& \\
\midrule
\texttt{\bfseries fc6}  & 411.0 MB  & 3\%  & 15.4 MB  && 2.70 MB & \\
\texttt{\bfseries fc7}  & 67.1 MB  & 4\%  & 3.4 MB  && 0.75 MB & \\
\texttt{\bfseries fc8}  & 16.4 MB  & 24\%  & 4.8 MB  && 0.83 MB & \\
\bf overall & 494.5 MB & 3.8\%  & 23.6 MB & \ \scriptsize(20.9 $\times$) & 4.28 MB & \ \scriptsize(115.6 $\times$) \\
\bottomrule
\end{tabular}
}
    \end{subtable}
\label{tab:fc-stat}
\end{table}

\subsubsection{Linearity of Accuracy Loss} We first experimentally demonstrate the approximate linearity of accuracy loss on \glspl{fclayer} based on AlexNet and VGG-16. 
We set different combinations of error bounds (i.e., within 0.1) for \glspl{fclayer} and compare our expected accuracy loss (based on Equation (\ref{eq:ADsum})) with the actual accuracy loss, as shown in Figure~\ref{fig:linearity}. Specifically, x-axis shows our expected accuracy loss, which is the sum of accuracy degradation in each \gls{fclayer}, and y-axis shows the actual accuracy loss. We can observe a clear linear relationship when the overall accuracy loss is lower than 2\%. Therefore, we can use this approximate linearity of accuracy loss to perform the following optimization of error bound configurations discussed in Section \ref{sec:fine-turning}.

\subsubsection{Compression Ratio}
We then present the experimental results of \gls{deepsz} in terms of compression ratio and compare with Deep Compression and Weightless.
\paragraph{LeNet-300-100 and LeNet-5 on MNIST} First, we evaluate  \gls{deepsz} on LeNet300-100 and LeNet-5 with the MNIST dataset \cite{lecun2010mnist}. LeNet-300-100 contains only three \glspl{fclayer} (i.e., \texttt{ip1}, \texttt{ip2}, and \texttt{ip3}). LeNet-5 contains three convolutional layers and two \glspl{fclayer} (i.e., \texttt{ip1} and \texttt{ip2}). The \glspl{fclayer} dominate: 100\% and 95.3\% of the overall sizes of LeNet-300-100 and LeNet-5, respectively, as shown in Table~\ref{tab:NeuralNetworks}.
\gls{deepsz} first prunes the network with the pruning ratios suggested by \cite{han2015deep} (as shown in Table \ref{tab:subtable_LeNet-300-100} and \ref{tab:subtable_LeNet-5}) and stores the pruned weights in the \glspl{dataArray} and \glspl{indexArray}. 
After the pruning step, LeNet-300-100 and LeNet-5 can be reduced by 9.7$\times$ and 9.8$\times$, respectively. Note that the compression ratio is slightly different from the pruning ratio because every nonzero pruned weight requires 40 bits instead of 32 bits, as discussed in Section~\ref{sec:design_02_pruning}. 
Then, \gls{deepsz} deploys the error bound assessment step to the pruned network and gets the feasible ranges of error bounds for \glspl{fclayer}. The feasible ranges are $[10^{-2}, 3 \times 10^{-2}]$, $[10^{-2}, 8 \times 10^{-2}]$, and $[10^{-2}, 2 \times 10^{-1}]$ for \texttt{ip1}, \texttt{ip2}, and \texttt{ip3} of LeNet-300-100, respectively, as shown in Figure~\ref{fig:subfigure_AccuracyDrop_LeNet-300-100}. The ranges are $[10^{-2}, 3 \times  10^{-2}]$ and $[10^{-2}, 9 \times 10^{-2}]$ for \texttt{ip1} and \texttt{ip2} of LeNet-5, respectively, as shown in Figure~\ref{fig:subfigure_AccuracyDrop_LeNet-5}.
 \gls{deepsz} then optimizes the configuration of the error bounds based on Algorithm~2. The final error bounds of \texttt{ip1}, \texttt{ip2}, and \texttt{ip3} of LeNet-300-100 are $2 \times 10^{-2}$, $3 \times 10^{-2}$, and $4 \times 10^{-2}$, respectively.
The final error bounds of \texttt{ip1} and \texttt{ip2} of LeNet-5 are $3 \times 10^{-2}$ and $8 \times 10^{-2}$, respectively. 
 \gls{deepsz} then adopts the best-fit lossless compressor---Zstandard---to compress the \glspl{indexArray}. As shown in Table~\ref{tab:subtable_LeNet-300-100} and \ref{tab:subtable_LeNet-5}, \gls{deepsz} can compress \glspl{fclayer} of LeNet-300-100 by 55.8$\times$ and the \glspl{fclayer} of LeNet-5 by 57.3$\times$ with \textit{no loss of inference accuracy}. 
We note that each \gls{fclayer} has a threshold of error bound, after which the inference accuracy begins to drop sharply.
This phenomenon is also true for other networks, such as AlexNet and VGG-16. It demonstrates that how to determine the proper error bounds for each \gls{fclayer} is a critical problem, for which \gls{deepsz} provides an efficient, fine-tuning solution (see Section \ref{sec:err-bound-assessment} and Section \ref{sec:fine-turning}).

\paragraph{AlexNet on ImageNet}
We next evaluate \gls{deepsz} on a much larger network, AlexNet, with the ImageNet dataset \cite{krizhevsky2012imagenet}.
AlexNet contains five convolutional layers and three \glspl{fclayer} (i.e., \texttt{fc6}, \texttt{fc7}, and \texttt{fc8}). The \glspl{fclayer} take up 96.1\% of the overall storage space, as shown in Table~\ref{tab:NeuralNetworks}.
After the pruning, the network can be reduced to 10.1\%, as shown in Table~\ref{tab:subtable_AlexNet}. After the second assessment and third optimization step, \gls{deepsz}  uses $7\times10^{-3}$, $7\times10^{-3}$, and $5\times10^{-3}$ as the error bound for \texttt{fc6}, \texttt{fc7}, and \texttt{fc8}, respectively, as shown in Figure \ref{fig:subfigure_AccuracyDrop_VGG-16}. 
\gls{deepsz} can compress AlexNet by 45.5$\times$ with only 0.13\% loss of top-1 accuracy, as shown in Table~\ref{tab:AccuracyDropAfterCompressing}. Note that the top-5 accuracy is not decreased but, rather, is increased by 0.18\%.
We can further set the expected inference accuracy loss to zero.  \gls{deepsz} then can compress AlexNet by 36.5$\times$ with no loss of inference accuracy  (the error bound of $2\times10^{-3}$ for all \texttt{fc6}, \texttt{fc7}, and \texttt{fc8}).



\paragraph{VGG-16 on ImageNet}
We now apply \gls{deepsz} on VGG-16, which  contains one large \gls{fclayer} (i.e., \texttt{fc6}) and two relatively small \glspl{fclayer} (i.e., \texttt{fc7} and \texttt{fc8}). 
The pruning ratios are set to relatively low values, leading to a much higher compression ratio after the pruning (i.e., 20.9$\times$), as shown in Table~\ref{tab:subtable_VGG-16}.  \gls{deepsz} then uses $10^{-2}$, $9\times10^{-3}$, and $5\times10^{-3}$ as the error bound for \texttt{fc6}, \texttt{fc7}, and \texttt{fc8}, respectively. By leveraging \gls{deepsz}, we can achieve a compression ratio of 115.6$\times$  with only 0.25\% loss of inference accuracy on VGG-16, as shown in Table~\ref{tab:AccuracyDropAfterCompressing}.
Similar to AlexNet, we can also set the expected inference accuracy loss to zero.  \gls{deepsz} then can compress VGG-16 by 92.7$\times$ with no loss of inference accuracy (with the error bound of $3\times10^{-3}$ for \texttt{fc6} and \texttt{fc7} and $10^{-3}$ for \texttt{fc8}). 


\begin{table}[t]
    \caption{Inference accuracy of \gls{deepsz} compressed LeNet-5, AlexNet, and VGG-16.}
    \vspace{-2mm}
    \label{tab:AccuracyDropAfterCompressing}
    \footnotesize\resizebox{\columnwidth}{!}{%
\begin{tabular}{@{}rrrrr@{}}
\toprule
\textbf{Neural Network} & 
    \begin{tabular}{@{}r@{}} \bfseries Top-1 \\\bfseries Accuracy \end{tabular} & 
    \begin{tabular}{@{}r@{}} \bfseries Top-5 \\\bfseries Accuracy \end{tabular} & 
    \begin{tabular}{@{}r@{}} \bfseries \glspl{fclayer}' \\\bfseries Size \end{tabular} & 
    \begin{tabular}{@{}r@{}} \bfseries Compress \\\bfseries Ratio \end{tabular} \\
\midrule
LeNet-300-100 original & 98.35\% &- & 1056 \makebox[1.5em][l]{KB} & \\
LeNet-300-100 \gls{deepsz} & 98.31\% &- & \bfseries 19.1 \makebox[1.5em][l]{KB} & 55.8$\times$ \\
\cmidrule{2-5}
LeNet-5 original & 
	99.13\% & 
	- & 
	1620 \makebox[1.5em][l]{KB} & \\
LeNet-5 \gls{deepsz} & 
	99.16\% & 
	- & 
	\bfseries 28.3 \makebox[1.5em][l]{KB} & 
	57.3 $\times$ \\
\cmidrule{2-5}
AlexNet original & 
	57.41\% & 
	80.40\% & 
	234.5 \makebox[1.5em][l]{MB} & 
	\\
AlexNet \gls{deepsz} & 
	57.28\% & 
	80.58\% & 
	\bfseries 5.15 \makebox[1.5em][l]{MB} & 
	45.5 $\times$ \\
\cmidrule{2-5}
VGG-16 original & 
	68.05\% & 
	88.34\% & 
	494.5 \makebox[1.5em][l]{MB} & 
	\\
VGG-16 \gls{deepsz} & 
	67.80\% & 
	88.20\% & 
	\bfseries 4.277 \makebox[1.5em][l]{MB} & 
	115.6 $\times$ \\
\bottomrule
\end{tabular}
}
\end{table}


In summary, \gls{deepsz} can compress the \glspl{fclayer} in the tested neural networks with  compression ratios of 57$\times$ to 116$\times$ while maintaining a loss of inference accuracy less than 0.3\% (within the user-set expected loss of 0.4\%), as shown in Table~\ref{tab:AccuracyDropAfterCompressing}. Note that the top-5 accuracy is usually not displayed on the LeNet-5 because its top-1 accuracy (i.e., > 99\%) is relatively high.
\gls{deepsz} can improve the overall compression ratio by 21\% to 43\%, compared with the second-best solution, as shown in Table~\ref{tab:CompressionRationCmp}. This table also illustrates that \gls{deepsz} can deliver a high compression ratio for each \gls{fclayer}. Even compared with the Weightless method, which  can  compress only one layer, \gls{deepsz} can still achieve a comparable compression ratio. We note that compression ratio is not available for some layers in Weightless, because (1) Weightless \cite{reagen2017weightless} does not provide their open source code and (2) the Weightless paper \cite{reagen2017weightless}  showed  evaluation results only for the largest two layers in LeNet-5 and VGG-16, without any results for AlexNet.
We also note that Deep Compression uses 5 bits per pruned weights, whereas \gls{deepsz} can compress the networks to 2.0 $\sim$ 3.3 bits per pruned weights. If we also set similar bit width for Deep Compression's quantization (i.e., the number of bits based on \gls{deepsz} compressed layers), the inference accuracy will drop sharply by 1.56\% for AlexNet and 2.81\% for VGG-16, as shown in Table~\ref{tab:AccuracyDegradation}. Note that the inference accuracy degradation is not available for Weightless for LeNet-5 and AlexNet, because Weightless does not provide these results in  \cite{reagen2017weightless} (the paper does show the inference accuracy degradation and encoding time overhead for VGG-16).



\begin{table}[t]
    \centering
    \caption{Comparison of compression ratios of different techniques on LeNet-300-100, LeNet-5, AlexNet, and VGG-16.}
    \label{tab:CompressionRationCmp}
    \footnotesize\begin{tabular}{@{}rrrrrr@{}}
\toprule
\multirow{3}{*}{\begin{tabular}{@{}r@{}} \bfseries Neural \\\bfseries Network \end{tabular}} 
    & \multirow{3}{*}{\begin{tabular}{@{}r@{}} \bfseries Layer \\ \end{tabular}} 
    & \multicolumn{4}{c}{\textbf{Compression Ratio}} \\ 
\cmidrule(l){3-6} 
    & 
    & \footnotesize Deep
    & \footnotesize Weight-
    & \multirow{2}{*}{\footnotesize \gls{deepsz}}
    & \footnotesize Improve- \\
    &
    & \footnotesize Compression
    & \footnotesize less
    &
    & \footnotesize ment \\
\midrule
\multirow{4}{*}{
    \begin{tabular}{@{}r@{}}
        \bfseries LeNet-\\\bfseries 300-100
    \end{tabular}
    } & 
    \texttt{ip1} & 43.1 & 60.1 & 61.81 & 1.43$\times$ \\
& \texttt{ip2} & 32.9 & 64.3 & 37.97 & 1.15$\times$ \\
& \texttt{ip3} & 7.9 & - & 5.6 & 0.71$\times$ \\
& \bfseries overall & \bfseries 41.0 & \bfseries 7.6 & \bfseries 55.77 & \bfseries 1.36$\times$ \\
\midrule
\multirow{3}{*}{\textbf{LeNet-5}} & 
    \texttt{ip1} & 40.8 & 74.2 &58.5 & 1.43$\times$ \\
& \texttt{ip2} & 16.3 & - &21.5 & 1.32$\times$ \\
& \bfseries overall & \bfseries 40.1 &\bfseries 39.0 &\bfseries 57.3 &\bfseries 1.43$\times$ \\
\midrule
\multirow{4}{*}{\textbf{AlexNet}} & 
    \texttt{fc6} & 41.8 & - & 54.4 & 1.30$\times$ \\
& \texttt{fc7} & 40.7 & - & 46.5 & 1.14$\times$ \\
& \texttt{fc8} & 17.1 & - & 17.5 & 1.02$\times$ \\
& \bfseries overall & \bfseries 37.7 & - & \bfseries 45.5 & \bfseries 1.21$\times$ \\
\midrule
\multirow{4}{*}{\textbf{VGG-16}} & 
    \texttt{fc6} & 119.0 & 157.0 & 152.1 & 1.28$\times$ \\
& \texttt{fc7} & 80.0 & 85.8 & 90.0 & 1.13$\times$ \\
& \texttt{fc8} & 19.1 & - & 19.8 & 1.04$\times$ \\
& \bfseries overall & \bfseries 95.8 & \bfseries 5.9 & \bfseries 115.6 & \bfseries 1.21$\times$ \\
\midrule
\end{tabular}

\end{table}

\begin{table}[]
    \caption{Inference accuracy degradation of different techniques based on comparable compression ratio.}
    \label{tab:AccuracyDegradation}
    \footnotesize\begin{tabular}{@{}rrrr@{}}
\toprule
\textbf{model} 
    & \textbf{\begin{tabular}[c]{@{}r@{}}quantization\\ (Deep Compression)\end{tabular}} 
        & \textbf{\begin{tabular}[c]{@{}r@{}}Bloomier Filter\\ (Weightless)\end{tabular}} 
            & \textbf{\begin{tabular}[c]{@{}r@{}}SZ\phantom{)}\\ (\gls{deepsz})\end{tabular}} \\
\midrule
LeNet-300-100	& 0.22\%  & -  & 0.12\%  \\
LeNet-5			& 0.30\%  & -  & $-$0.03\%\footnote{The accuracy is slightly increased by 0.03\% in this case.} \\
AlexNet		& 1.56\%  & -  & 0.13\%  \\
VGG-16			& 2.81\%  & $>$3.0\%  & 0.25\%  \\
\bottomrule
\end{tabular}
\end{table}



\subsubsection{Performance Evaluation}
\begin{figure}[]
    \centering
    \begin{subfigure}{\linewidth}
        \includegraphics[width=0.9\linewidth]{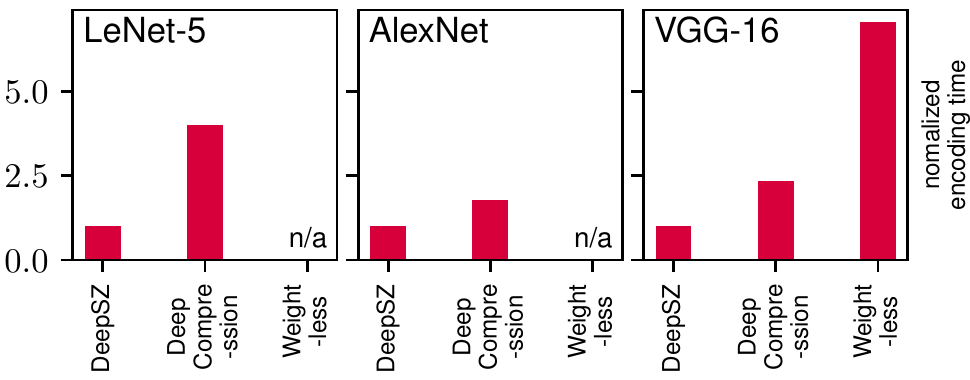}
        \caption{Normalized encoding time with different solutions on GPUs.}
        \label{fig:subfigure_TimeBreakdownComp}
    \end{subfigure}
    \begin{subfigure}{\linewidth}
    \vspace{4mm}
        \includegraphics[width=0.9\linewidth]{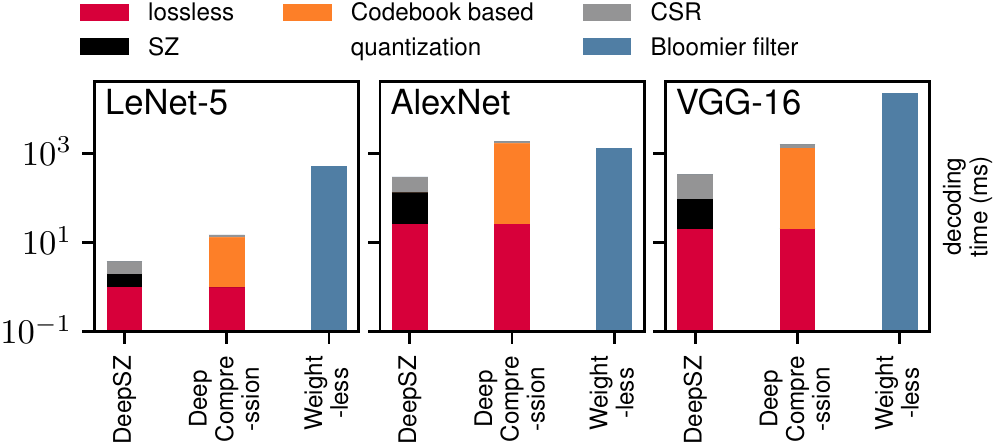}
        \caption{Breakdown of decoding time with different solutions on CPU.}
        \label{fig:subfigure_TimeBreakdownDecomp}
    \end{subfigure}
    
    \caption{Time breakdown of encoding and decoding with different lossy compression techniques.}\label{fig:TimeBreakdown}
\end{figure}

As discussed in Section~\ref{sec:comparison}, \gls{deepsz} is faster than the other methods theoretically in terms of both encoding and decoding. We now present the time overhead of \gls{deepsz} on the four neural networks, as shown in Figure~\ref{fig:TimeBreakdown}. The figure illustrates that \gls{deepsz} has lower encoding and decoding time overheads than do Deep Compression and Weightless. We note that the time results of LeNet-300-100 are almost identical to those of LeNet-5; hence, because of space limitations, we  present the time overheads only for LeNet-5. 

We investigated the times of the last three steps (i.e.,  spent mainly in the time of compression, decompression, and tests) for \gls{deepsz}'s encoding on GPUs. We do not include the pruning time because all  three methods have the same pruning process and the time overheads are the same. 
Figure~\ref{fig:subfigure_TimeBreakdownComp} shows the encoding time with the three  solutions. We normalize the other two compression methods compared with \gls{deepsz} in Figure~\ref{fig:subfigure_TimeBreakdownComp}, because compared with AlexNet and VGG-16, LeNet-5 features much smaller encoding time. Specifically, \gls{deepsz} takes <1 min, 8 min, and 16 min on encoding LeNet-5, AlexNet, and VGG-16, respectively. Deep Compression takes 4 min, 14 min, and 38 min on encoding LeNet-5, AlexNet, and VGG-16, respectively. 
Due to lack of source code, we estimate the encoding time of Weightless based on the number of epochs (for retraining) shown in the paper and the time of one epoch based on our experimental platform. Weightless takes about 113 min on encoding VGG-16; 
again, Weightless does not present the encoding time (i.e., the number of epochs) of LeNet-5 or AlexNet in the paper. \gls{deepsz} can improve the encoding performance by 1.8$\times$ to 4.0$\times$ compared with the second-best solution. We note that for the Deep Compression and Weightless methods, it is difficult to determine the initial parameters of the solver in order to retrain the network. It could take much longer time than the optimal performance overhead if users are not familiar with the characteristics of the network.

We also investigated the times of lossless decompression, SZ lossy decompression, and sparse matrix reconstruction for \gls{deepsz}'s decoding on CPU. As we can see in Figure~\ref{fig:subfigure_TimeBreakdownDecomp},  \gls{deepsz} outperforms the second-best solution by 4.5$\times$ to 6.2$\times$ for decoding. Specifically, \gls{deepsz} takes 2.7 ms, 296 ms, and 341 ms on decoding LeNet-5, AlexNet, and VGG-16, respectively; Deep Compression takes 13.9 ms, 1,832 ms, and 1,565 ms on decoding LeNet-5, AlexNet, and VGG-16, respectively; and Weightless takes 520 ms, 1,300 ms, and 22,800 ms on decoding LeNet-5, AlexNet, and VGG-16, respectively, as shown in the paper \footnote{The paper evaluated its decoding time on an Intel Core i7-6700K Processor, which has similar processing power to our processor.}. More specifically, for example, \gls{deepsz} spends 26 ms in lossless decompression, 108 ms in SZ lossy decompression, and 162 ms in reconstructing the sparse matrix on AlexNet. As a comparison, the time for one forward pass with 50 images per batch takes 1,100 ms on AlexNet. This demonstrates that the time overhead of \gls{deepsz}'s decoding is comparatively low compared with typical forward pass. Therefore, once the network is needed for inference, \gls{deepsz} can quickly decompress the compressed data and reconstruct the network  without much delay. Note that the decoding time of Weightless relies on the number of nonpruned weights, whereas the decoding times of \gls{deepsz} and Deep Compression depend on the number of pruned weights. This difference can explain the following two observations. (1) \gls{deepsz} and Deep Compression have similar decoding time on AlexNet and VGG-16 because they have similar numbers of pruned weights (i.e., 6.5 million for AlexNet and 5.8 million for VGG-16). (2) Weightless spends more time on VGG-16 than AlexNet for decoding because the largest \gls{fclayer} of VGG-16 (i.e., \texttt{fc6} of 25,088$\times$4,096) is much larger than that of AlexNet (i.e., \texttt{fc6} of 9,216$\times$4,096).

\section{Related Work}
\label{sec:related}

Most of neural networks have significant redundancy in their parameters according to a well-known research study \cite{denil2013predicting}. Such redundant information may cause significant waste of computation, memory, and storage resources. In general,  two types of methods have been proposed to resolve this issue: (1) modifying the network structures to reduce the complexity of parameters and (2) compressing a trained network by removing redundant information. 

Modifying the structures of networks by adopting specialized structure or loss function can reduce the memory footprint while training the larger-scale networks with the same resources.
For example, Vanhoucke et al. \cite{vanhoucke2011improving} exploited a fixed-point representation of activations with 8-bit integer rather than 32-bit floating point. Denton et al. \cite{denton2014exploiting} proposed using low-rank tensor approximations to reduce the number of parameters by up to a factor of 13 for a single layer while keeping the inference accuracy loss of 1\% compared with the original network. Arora et al. \cite{arora2014provable} theoretically studied using random-like sparse networks with +1/0/-1 weights for interesting properties.
Chen et al. \cite{chen2015compressing} proposed a network architecture, named HashedNets, that uses a low-cost hash function to randomly group connection weights into hash buckets, such that all connections within the same hash bucket share a single value. 

Compressing neural networks is an alternative strategy to reduce the model size. For example, Gong et al. \cite{gong2014compressing} compressed \glspl{fclayer} by using vector quantization, which achieved a compression ratio of 24 with 1\% inference accuracy loss.
Recently, two state-of-the-art works \cite{han2015deep, reagen2017weightless} have been designed for compressing the network with high compression ratio and inference accuracy.
Han et al. proposed a three-step approach, named Deep Compression, that contains pruning, quantization, and encoding. Deep Compression, however, may degrade the inference accuracy significantly in the course of each forward-propagation because of its vector quantization design, such that the network has to to be retrained over and over again in order to reach the target inference accuracy, thus resulting in  a high execution time overhead.
Reagen et al. proposed a lossy compression method, named Weightless, by adopting a Bloomier filter to compress the weights lossily. For encoding, the Bloomier filter needs to construct a hash table, which is $\mathcal{O}(n\log n)$ in time complexity; for decoding, in order to decompress one value, the Bloomier filter typically needs to calculate  four hashing functions. The time complexity is $\mathcal{O}(n)$ for the best case but $\mathcal{O}(n^2)$ for the worst case. Here $n$ is the number of values for encoding/decoding.
Therefore, the Weightless method suffers from a relatively high time overhead because of the expensive Bloomier filter. Moreover, it was  applied to only one \gls{fclayer} instead of the whole neural network.
Tung et al. \cite{tung2018clip} proposed a method named CLIP-Q that uses weight pruning and quantization, which is similar to Deep Compression. Unlike Deep Compression that separates pruning and quantization, CLIP-Q combines them at the same step in a single framework and can be performed in parallel with network fine-tuning. Moreover, it adopts much higher pruning ratio than Deep Compression in order to achieve higher overall compression ratio.
In this paper, we mainly compare our proposed \gls{deepsz} with both Deep Compression and Weightless approaches comprehensively.

Unlike the first type of method that requires modification of the network structure and full retraining, the second type of method is more general and efficient. Therefore, we focus on compressing well-trained neural networks without modifying the network structure for high reduction ratio and inference accuracy. 

\section{Conclusion and Future Work}
\label{sec:conclusion}

In this paper, we propose a novel lossy compression framework, called \gls{deepsz}, for effectively compressing sparse weights in deep neural networks. Unlike  traditional methods, \gls{deepsz} can avoid the costly retraining process after compression, leading to a significant performance improvement in encoding DNNs.
We develop a series of approaches to efficiently determine the best-fit error bound for each layer in the network, maximizing the overall compression ratio with user acceptable loss of inference accuracy. Experimental results based on the tested neural networks show that \gls{deepsz} can achieve  compression ratios of up to 116$\times$ and can  outperform the second-best approach by up to 1.43$\times$. Our experiments with four Nvidia Tesla V100 GPUs demonstrate that \gls{deepsz} can obtain 1.8$\times$ to 4.0$\times$ performance improvement in encoding compared with the previous state-of-the-art. \gls{deepsz} can improve the decoding performance by 4.5$\times$ to 6.2$\times$ compared with the second-best solution. \gls{deepsz} also can provide high flexibility to balance the compression ratio and inference accuracy. We plan to first evaluate our proposed \gls{deepsz} on more neural network architectures. 
We also will further improve the SZ compression algorithm to achieve a higher reduction ratio in compressing DNNs. Moreover, we hope to use \gls{deepsz} for improving GPU memory utilization.

\section*{Acknowledgments}
\footnotesize
This research was supported by the Exascale Computing Project (ECP), Project Number: 17-SC-20-SC, a collaborative effort of two DOE organizations -- the Office of Science and the National Nuclear Security Administration, responsible for the planning and preparation of a capable exascale ecosystem, including software, applications, hardware, advanced system engineering and early testbed platforms, to support the nation's exascale computing imperative. This material was based upon work supported by the U.S. Department of Energy, Office of Science, under contract DE-AC02-06CH11357, and also supported by the National Science Foundation under Grant No. 1619253. We gratefully acknowledge the support from Alabama Water Institute (AWI), Remote Sensing Center (RSC), and Center for Complex Hydrosystems Research (CCHR).

\bibliographystyle{ACM-Reference-Format}
\bibliography{refs}

\end{document}